\documentclass[lettersize,journal]{IEEEtran}
\usepackage{cite}
\usepackage{amsmath,amssymb,amsfonts}
\usepackage{graphicx}
\usepackage{textcomp}
\usepackage{xcolor}
\usepackage{multirow}
\usepackage{makecell}
\usepackage{booktabs}
\usepackage{comment}
\usepackage{algorithm}
\usepackage{algpseudocode}
\usepackage{subcaption}
\usepackage{url}
\usepackage{hyperref}
\hypersetup{
    colorlinks= true,
    linkcolor= black,
    filecolor= black,
    citecolor = black, 
    urlcolor= blue,
}
\def\BibTeX{{\rm B\kern-.05em{\sc i\kern-.025em b}\kern-.08em
    T\kern-.1667em\lower.7ex\hbox{E}\kern-.125emX}}

\begin{document}

\title{CNN-TFT explained by SHAP with multi-head attention weights for time series forecasting}

\author{Stefano F. Stefenon, Jo\~ao P. Matos-Carvalho, Valderi R. Q. Leithardt,~\IEEEmembership{Senior Member, IEEE}, \\and Kin-Choong Yow,~\IEEEmembership{Senior Member, IEEE}

\thanks{
This work was supported in part by the Natural Sciences and Engineering Research Council of Canada (NSERC) under Grant DDG-2024-00035, and in part by the Cette recherche a \'et\'e financ\'ee par le Conseil de recherches en sciences naturelles et en g\'enie du Canada (CRSNG) under Grant DDG-2024-00035. This work was also supported by the Portuguese Agency FCT (Funda\c{c}\~ao para a Ci\^encia e Tecnologia), under Grant LASIGE Research Unit, ref. UID/00408/2025.
		
S. F. Stefenon is with the Lisbon School of Engineering (ISEL), Polytechnic University of Lisbon, Lisbon 1959-007, Portugal, and Faculty of Engineering and Applied Sciences, University of Regina, Saskatchewan, S4S 0A2, Canada. (e-mail: stefano.stefenon@isel.pt) 
        
K.-C. Yow is with the Faculty of Engineering and Applied Sciences, University of Regina, Saskatchewan, S4S 0A2, Canada.

J. P. Matos-Carvalho is with LASIGE, Departamento de Inform\'atica, Faculdade de Ci\^encias, Universidade de Lisboa, 1749--016 Lisboa, Portugal, and Center of Technology and Systems (UNINOVA-CTS) and LASI, 2829-516 Caparica, Portugal.

V. R. Q. Leithardt is with Instituto Universit\'ario de Lisboa, (ISCTE-IUL), ISTAR, 1649-026, Lisboa, Portugal.

}
\thanks{Manuscript received April 19, 2021; revised August 16, 2025.}}

\markboth{Journal of \LaTeX\ Class Files,~Vol.~14, No.~8, August~2025}%
{Shell \MakeLowercase{\textit{et al.}}: A Sample Article Using IEEEtran.cls for IEEE Journals}


\maketitle

\begin{abstract}
Convolutional neural networks (CNNs) and transformer architectures offer strengths for modeling temporal data: CNNs excel at capturing local patterns and translational invariances, while transformers effectively model long-range dependencies via self-attention. This paper proposes a hybrid architecture integrating convolutional feature extraction with a temporal fusion transformer (TFT) backbone to enhance multivariate time series forecasting. The CNN module first applies a hierarchy of one-dimensional convolutional layers to distill salient local patterns from raw input sequences, reducing noise and dimensionality. The resulting feature maps are then fed into the TFT, which applies multi-head attention to capture both short- and long-term dependencies and to weigh relevant covariates adaptively. We evaluate the CNN-TFT on a hydroelectric natural flow time series dataset. Experimental results demonstrate that CNN-TFT outperforms well-established deep learning models, with a mean absolute percentage error of up to 2.2\%. The explainability of the model is obtained by a proposed Shapley additive explanations with multi-head attention weights (SHAP-MHAW). Our novel architecture, named CNN-TFT-SHAP-MHAW, is promising for applications requiring high-fidelity, multivariate time series forecasts, being available for future analysis at \url{https://github.com/SFStefenon/CNN-TFT-SHAP-MHAW}.
\end{abstract}


\section{Introduction}
\IEEEPARstart{T}{his} series forecasting, especially in the field of energy, has been widely explored, and predictive models have been proposed that show promising results for application in real-world situations \cite{tian2025sdvs}. Within this field, hybrid models stand out for using the advantages of several methods combined to provide a structure that outperforms state-of-the-art models. One combination of models that has been successfully applied is the convolutional neural network (CNN) with long short-term memory (LSTM) using the attention mechanism \cite{wan2023short}, or called attention-based CNN-LSTM \cite{mou2021driver}. 

CNNs automatically extract hierarchical local patterns and temporal dependencies through convolutional filters, which reduces the need for manual feature engineering \cite{zhao2025short}. By sliding kernels across the input sequence, CNNs efficiently capture multi-scale trends, seasonality, and cyclical patterns while leveraging parameter sharing to reduce computational complexity. Their translation invariance allows them to recognize recurring motifs regardless of position, and their parallel processing capability accelerates training \cite{lu2021cnn}. 

The attention mechanism dynamically weighs the influence of historical observations, enabling the model to focus on relevant time steps while ignoring noise \cite{narhi2023attention}. This adaptability enhances the capture of long-range dependencies, which is essential in sequences where distant past events influence future values \cite{gao2021interpretable}. 

Temporal fusion transformer may outperform LSTM for time series forecasting due to its ability to model complex temporal relationships and handle heterogeneous inputs (static, known future, and observed time-varying features) \cite{STEFENON2024109876}. Unlike LSTMs, which process sequences sequentially and struggle with long-term dependencies, the temporal fusion transformer (TFT) employs multi-head self-attention to capture both short- and long-range patterns dynamically, improving accuracy in noisy, multi-variate datasets. 

Based on the promising results of the TFT model compared to the LSTM model, considering the advantages of using CNN for feature extraction \cite{10443012}, and attention mechanism to enhance predictions, this paper proposes a new model that combines the CNN and the TFT. The proposed model uses causal 1D convolutional blocks (\texttt{Conv1D}) in the encoder, leveraging CNNs' efficiency and parallelism on local patterns. This modification preserves the rest of the TFT pipeline while swapping out the recurrent core for a convolutional one.

Shapley additive explanations (SHAP) can help provide explanations for model predictions, helping to build trust and transparency in complex models \cite{10443379}. By fairly distributing the contribution of each feature based on cooperative game theory (Shapley values), SHAP offers both global insights into feature importance and local explanations for individual predictions \cite{10402107}. SHAP's additive nature allows interpretation of how each feature influences the outcome, which is especially valuable for model validation \cite{10429777}.


To improve models' interpretability, this paper proposes a SHAP with multi-head attention weights (MHAW). Integrating SHAP values with attention offers an advancement in model explainability by combining two complementary perspectives: attention weights, which reveal where the model focuses during prediction, and SHAP values, which quantify the causal contribution of each input feature.

Our proposed model, called CNN-TFT-SHAP-MHAW, has the following contributions:

\begin{itemize}
    \item We propose a new CNN-TFT-SHAP-MHAW model written from scratch that combines the advantages of both models to create an innovative hybrid structure for time series forecasting.

    \item The hybrid structure combines CNN and attention outputs for enhanced feature extraction, ensuring that the most important features are used for signal prediction. 

    \item Multi-head attention learns temporal relationships, 
    capturing long-range dependencies, helping in the identification of relationships between distant time steps.

    \item Our model uses causal padding in convolutional layers to ensure predictions only depend on past/current inputs (no data leakage from future values).
    \item By merging attention weights with SHAP values, this approach creates a more interpretable and trustworthy influence map that reveals both what the model focuses on and why, enhancing explainability to support better decision-making.
\end{itemize}

The remainder of this paper is organized as follows: Section \ref{2a} presents a discussion of related works. 
Section \ref{2} presents the architecture of the proposed CNN-TFT-SHAP-MHAW model. In Section \ref{3}, the dataset and experiment setup are given. In Section \ref{4}, results and discussion are presented, and finally, in Section \ref{5}, a conclusion and future directions of research are drawn. 

\section{Related Works} \label{2a}

Recent work in long-term photovoltaic (PV) load forecasting has primarily relied on classical approaches such as regression-based models \cite{10566494}, which depend heavily on extensive historical data covering electricity load, weather, economic, and demographic variables \cite{9785904}. While these methods have seen widespread use, they often struggle with capturing long-term trends, handling complex variable interactions, and maintaining accuracy over extended forecasting horizons \cite{stefenon2024hypertuned}. 

To address these challenges, recent studies have explored hybrid deep learning (DL) models \cite{abou2023coa}, such as combinations of CNNs and LSTM networks (CNN-LSTM), which have demonstrated improved performance in capturing temporal and spatial patterns in data. These advanced architectures offer enhanced adaptability to trend shifts and better scalability for high-dimensional input data, setting a new benchmark for forecasting accuracy and robustness \cite{10916641}.

Aslam et al. \cite{9496627} proved that the two-stage attention mechanisms over LSTM networks have demonstrated superior performance in capturing complex temporal dependencies and emphasizing critical input features such as solar radiation, temperature, and humidity. The integration of Bayesian optimization for hyperparameter tuning has led to further improvements in prediction accuracy. Comparative evaluations with existing models, including CNN-LSTM, LSTM-attention, and other single attention frameworks, have shown that these advanced models offer enhanced forecasting skill and lower error metrics, underscoring their robustness and practicality for real-world PV power forecasting applications.

Clouds significantly influence surface irradiance, directly impacting PV output, and accurate real-time forecasting requires effectively linking sky conditions to solar irradiance levels. To this end, hybrid DL models combining convolutional autoencoders for feature extraction with clustering algorithms have been introduced to process sky image data and capture complex spatial patterns of cloud cover. These models then map the extracted features to surface irradiance using various DL architectures \cite{9054985}.

Expanding the scope of solar PV forecasting to cross-regional and international applications, recent research has explored the use of transfer learning to overcome limitations posed by scarce historical data at newly established PV sites \cite{10786977}. 
By integrating external weather variables and satellite imagery, the approach enables the model to generalize well across spatial contexts, achieving substantial gains in adjusted coefficient of determination values compared to conventional models like LSTM and CNN-LSTM. 

In arid regions such as Hail, Saudi Arabia, where climatic conditions are extreme and variable, DL architectures have shown promising capabilities for one-day-ahead irradiation forecasting. Models such as LSTM, bidirectional LSTM, gated recurrent unit (GRU), bidirectional GRU, and their hybrid combinations, such as CNN-LSTM and CNN-BiLSTM, have been evaluated by Boubaker et al. \cite{9363177} using a 20-year daily irradiation dataset. Based on these analyses, they were able to obtain a correlation coefficient of up to 96\%. Their findings highlight the potential of a hybrid DL model applied for enhancing solar resource forecasting in data-constrained or meteorologically harsh environments.

According to \cite{10876149}, the CNN with wavelet neural networks (WNN) masked multi-head attention (CNN-WNN-MMHA) is a promising solution by integrating spatial feature extraction through CNNs, multi-scale frequency analysis via wavelet neural networks WNN, and temporal sequence modeling with MMHA. This hybrid architecture allows for a comprehensive understanding of irradiance behavior under diverse climatic conditions. Their model achieved a 79\% reduction in mean absolute percentage error (MAPE), indicating improved prediction accuracy. The results also showed better generalization across varying weather conditions, with consistent performance in capturing both short-term fluctuations and long-term patterns in irradiance data.

Explainable artificial intelligence is an analysis trend that is being explored to improve the interpretability of model results  \cite{10879535}. In time series contexts, SHAP helps uncover how past values, time-based features (such as lagged variables, trends, or seasonal indicators), and exogenous inputs influence forecasted values \cite{10669041}. By assigning an influence value to each feature at each time step, SHAP enables the interpretation of complex models, revealing which inputs most significantly drive forecasts \cite{zuege2025wind}. 

As shown in \cite{9496627}, hypertuning is promising for adjusting the model and using an optimized structure. As outlined in \cite{9054985}, the non-linearities caused by irradiance variation result in challenges in prediction, showing promise for the use of hybrid models. As discussed in \cite{10111057}, the use of CNN for feature extraction improves the model's predictive capacity. 
Based on these previous research findings, this paper proposes a model created from scratch, called CNN-TFT-SHAP-MHAW, optimized by Bayesian optimization for PV forecasting.

\section{CNN-TFT-SHAP-MHAW Architecture} \label{2}

\begin{figure}[htb!]
	\centering
	\setkeys{Gin}{width=.5\textwidth}
	{\includegraphics{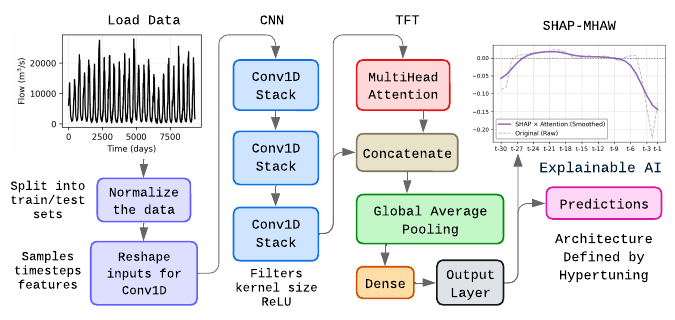}}
	\caption{ \label{f-flow}Summarized architecture of the proposed CNN-TFT-SHAP-MHAW.}
\end{figure}

Figure \ref{f-flow} presents the architecture of the proposed CNN-TFT-SHAP-MHAW. The proposed model is a hybrid DL architecture designed for time series forecasting, combining the strengths of CNNs and transformer-based attention mechanisms. The model begins with causal convolutional layers to efficiently extract local temporal features while preserving the autoregressive structure of time series data (presented in subsection \ref{2b}). 

These representations are then processed by a multi-head self-attention mechanism, which captures long-range dependencies and temporal relationships across the input window (see subsection \ref{2c}). By fusing the outputs of both modules, the model leverages both short-term patterns and global context to make better predictions (see subsection \ref{2d}). To ensure that the best hyperparameters are used, Bayesian optimization is employed; this method is explained in subsection \ref{hippp}. The interpretability, considering the proposed SHAP-MHAW, is explained in subsection \ref{shap}.

\subsection{Causal Convolutional Layers} \label{2b}

To capture local temporal dynamics in the input time series, the model employs a stack of 1D causal convolutional layers (\texttt{Conv1D}). Given an input sequence $X \in \mathbb{R}^{w \times 1}$, the first convolutional layer applies a 1D causal convolution with $f$ filters and kernel size $k$:
\begin{equation}
H^{(1)} = \text{ReLU}\left(\text{Conv1D}_{\text{causal}}^{(f,k)}(X)\right),
\end{equation}
where $\text{Conv1D}_{\text{causal}}^{(f,k)}$ indicates a convolution with $f$ filters of width $k$ using causal padding. Causal padding ensures that the output at time $t$ only depends on inputs from time $t$ and earlier, preserving the autoregressive structure required for forecasting:
\begin{equation}
h^{(1)}_t = \sigma\left( \sum_{i=0}^{2} W_i \cdot x_{t-i} + b \right).
\end{equation}

A second identical causal convolution is then applied:
\begin{equation}
H^{(2)} = \text{ReLU}\left(\text{Conv1D}_{\text{causal}}^{(f,k)}(H^{(1)})\right),
\end{equation}
resulting in deeper hierarchical features that encode increasingly abstract temporal patterns. By stacking causal convolutional layers with kernel size $k$, the model effectively captures dependencies across a short historical context \cite{buratto2024wavelet}.

Each convolutional layer learns distinct filters that act as pattern detectors over local subsequences. The rectified linear activation function (ReLU) introduces non-linearity, enabling the network to model complex, piecewise linear transformations of the input.
Stacking multiple causal convolutional layers allows the network to model longer temporal dependencies while preserving an autoregressive structure \cite{8947933}.

This CNN block acts as a temporal feature extractor, transforming the raw time series into a higher-dimensional representation that emphasizes recent local structures, which is then passed to the attention mechanism for global reasoning.

\subsection{Multi-Head Self-Attention} \label{2c}

Following the convolutional layers, we incorporate a multi-head self-attention mechanism to model global temporal dependencies. Let $H \in \mathbb{R}^{w' \times d}$ be the output of the CNN stack. Self-attention computes attention scores between all time steps:
\begin{equation}
\text{Attention}(Q, K, V) = \text{softmax}\left(\frac{QK^\top}{\sqrt{d_k}}\right)V,
\end{equation}
where $Q = HW^Q$, $K = HW^K$, and $V = HW^V$ are linear projections of $H$, and $W^Q, W^K, W^V \in \mathbb{R}^{d \times d_k}$ are learnable weights \cite{yu2024self}.

In multi-head attention (MHA), this operation is performed in parallel over $h$ different heads:
\begin{equation}
\text{MHA}(H) = \text{Concat}(\text{head}_1, \ldots, \text{head}_h)W^O,
\end{equation}
where each head is a separate attention block with its own parameters, and $W^O$ is the final output projection matrix \cite{zhang2023integrated}.

\subsection{Feature Fusion and Prediction} \label{2d}

The CNN outputs $H_\text{CNN} \in \mathbb{R}^{w' \times d}$ and the attention outputs $H_\text{ATT} \in \mathbb{R}^{w' \times d'}$ are concatenated to form a fused representation:
\begin{equation}
H_\text{fused} = \text{Concat}(H_\text{CNN}, H_\text{ATT}) \in \mathbb{R}^{w' \times (d + d')}.
\end{equation}
To reduce dimensionality and aggregate temporal information, a global average pooling layer is applied:
\begin{equation}
z = \frac{1}{w'} \sum_{t=1}^{w'} H_\text{fused}^{(t)} \in \mathbb{R}^{d + d'}.
\end{equation}
Finally, a dense output layer maps the pooled vector to the target prediction:
\begin{equation}
\hat{y}_{T+1} = W_{\text{out}} z + b_{\text{out}}, \quad W_{\text{out}} \in \mathbb{R}^{1 \times (d + d')},\ b_{\text{out}} \in \mathbb{R}.
\end{equation}



The model's architecture is promising for time series forecasting because it combines the strengths of convolutional and transformer-based models. The causal convolutional layers efficiently capture short-term temporal patterns and local dependencies with low computational cost, while ensuring that predictions respect the chronological order of data. 

On top of this, the multi-head self-attention mechanism from the transformer framework enables the model to learn long-range dependencies and temporal relationships that may span across the entire input window, something CNNs alone struggle with. 
By concatenating both feature representations and using global pooling to summarize temporal dynamics, the model balances efficiency, making it well-suited for complex forecasting tasks where both local trends and global context are important.

\subsection{Hypertuning by Bayesian Optimization} \label{hippp}



In this paper, Bayesian optimization is used for hyperparameter tuning, where 
we wish to solve the optimization problem
\begin{align}
\mathbf{x}^* \;=\;\arg\max_{\mathbf{x}\in\mathcal{X}}\,f(\mathbf{x}),
\end{align}
where \(\mathcal{X}\subset\mathbb{R}^d\) is the hyperparameter domain and \(f\) is expensive to evaluate \cite{aghaabbasi2023hyperparameter}.
We place a Gaussian process ($\mathcal{GP}$) prior on \(f\):
\begin{align}
f(\mathbf{x}) \;\sim\;\mathcal{GP}\bigl(m(\mathbf{x}),\,k(\mathbf{x},\mathbf{x}')\bigr),
\end{align}
where \(m\colon\mathcal{X}\to\mathbb{R}\) is the mean function (often zero) and \(k\colon\mathcal{X}\times\mathcal{X}\to\mathbb{R}\) is the covariance kernel (e.g.\ squared exponential)
\begin{align}
k(\mathbf{x},\mathbf{x}') \;=\;\sigma_f^2\exp\!\Bigl(-\tfrac{1}{2}(\mathbf{x}-\mathbf{x}')^\top\Lambda^{-1}(\mathbf{x}-\mathbf{x}')\Bigr).
\end{align}

Given observed data \(\mathcal{D}_n=\{(\mathbf{x}_i,y_i)\}_{i=1}^n\), where \(y_i=f(\mathbf{x}_i)+\varepsilon_i\) with \(\varepsilon_i\sim\mathcal{N}(0,\sigma_n^2)\), the $\mathcal{GP}$ posterior at a candidate \(\mathbf{x}\) is Gaussian:
\begin{align}
\begin{aligned}
\mu_n(\mathbf{x}) &= k(\mathbf{x},X)\bigl[K + \sigma_n^2 I\bigr]^{-1}\mathbf{y},\\
\sigma_n^2(\mathbf{x}) &= k(\mathbf{x},\mathbf{x}) - k(\mathbf{x},X)\bigl[K + \sigma_n^2 I\bigr]^{-1}k(X,\mathbf{x}),
\end{aligned}
\end{align}
where \(X=[\mathbf{x}_1,\dots,\mathbf{x}_n]^\top\), \(\mathbf{y}=[y_1,\dots,y_n]^\top\), and \(K_{ij}=k(\mathbf{x}_i,\mathbf{x}_j)\).

An acquisition function \(\alpha_n(\mathbf{x})\) balances exploration and exploitation. Common choices include, where the
Expected improvement (EI) is:
\begin{align}
\begin{aligned}
\alpha_{\mathrm{EI}}(\mathbf{x}) \;=\;\mathbb{E}_{f\sim\mathcal{N}(\mu_n,\sigma_n^2)}\bigl[\max(f(\mathbf{x}) - f^+,\;0)\bigr]
= \\
\sigma_n(\mathbf{x})\bigl(u\,\Phi(u) + \varphi(u)\bigr),
\end{aligned}
\end{align}
where
\begin{align}
u = \frac{\mu_n(\mathbf{x}) - f^+ - \xi}{\sigma_n(\mathbf{x})},
\quad
f^+ = \max_{i\le n} y_i,
\end{align}
\(\Phi\) and \(\varphi\) denote the standard normal cumulative distribution function and probability density function, and \(\xi\ge0\) is an optional exploration parameter \cite{9037259}.

Bayesian optimization navigates expensive-to-evaluate hyperparameter spaces by modeling the unknown objective with a $\mathcal{GP}$ surrogate,
quantifying uncertainty to trade off exploration and exploitation via acquisition functions, and iteratively proposing new hyperparameter settings that maximize information gain or potential improvement \cite{10583855}. 

\subsection{Explainable Results} \label{shap}

To improve the interpretability of our model applied to univariate time series forecasting, two complementary explainability techniques were employed: SHAP with multi-head attention visualization. Algorithm~\ref{algo:shap_attention} presents a pseudo-code that depicts these descriptions, where $x$ is a sample of $X_{test}$.

\begin{algorithm}
\caption{Explainability pipeline with SHAP-MHAW.}
\label{algo:shap_attention}
\begin{algorithmic}[1]
\Require Trained the model $f$, input series $X \in \mathbb{R}^{T \times 1}$, window size $w$
\Ensure Smoothed influence map $\tilde{c}$

\State Select background set $B \subset X_{\text{train}}$
\State Select test input $x \in X_{\text{test}}$ with shape $(w, 1)$

\State Compute MHAW:
\Statex \hspace{1em} $A \gets \texttt{attention\_model}(x)$ \Comment{Shape: $(H, w, w)$}
\State Compute mean attention:
\Statex \hspace{1em} $a \gets \text{mean\_attention}(A)$ \Comment{Shape: $(w,)$}

\State Compute SHAP values:
\Statex \hspace{1em} $s \gets \texttt{SHAP}(f, x, B)$ \Comment{Shape: $(w,)$}

\State Combine SHAP and Attention:
\Statex \hspace{1em} $c \gets s \odot a$ \Comment{Element-wise multiplication}

\State Apply Gaussian smoothing:
\Statex \hspace{1em} $\tilde{c} \gets \texttt{GaussianFilter1D}(c, \sigma)$
\State \Return $\tilde{c}$
\end{algorithmic}
\end{algorithm}

SHAP is used here to quantify the contribution of each time step within the input window to the model's prediction. For each prediction, SHAP assigns an influence value to every lag.
Considering that lag refers to a previous time step's value of a time-dependent variable used as an input to predict future values. Positive SHAP values indicate that a given input step increased the predicted value, while negative values represent steps that pushed the forecast downward. The absolute magnitude of the SHAP values reflects the impact of each lag, regardless of direction \cite{9225146}.

To synthesize both perspectives, the SHAP values were element-wise multiplied by the corresponding multi-head attention weights (MHAW), resulting in a combined influence map (SHAP-MHAW).
This map highlights input time steps that were both attended to and had a strong causal influence on the forecast. 

\section{Experiment Setup and Dataset}
\label{3}

The NVIDIA RTX 5000 graphics processing unit (GPU), 16 GB of double data rate 6 synchronous dynamic random-access memory, was employed in this study to support computationally intensive tasks. The GPU was integrated into a Linux-based cluster and interfaced using compute unified device architecture to ensure compatibility with the DL and simulation frameworks utilized in this research.

The proposed CNN-TFT-SHAP-MHAW was written using the Python language. For model assessment, the root mean squared error (RMSE), mean absolute error (MAE), MAPE, and mean squared logarithmic error (MSLE) \cite{liu2022effective} were considered. 

\begin{figure}[htb!]
	\centering
	\setkeys{Gin}{width=0.49\textwidth}
	{\includegraphics{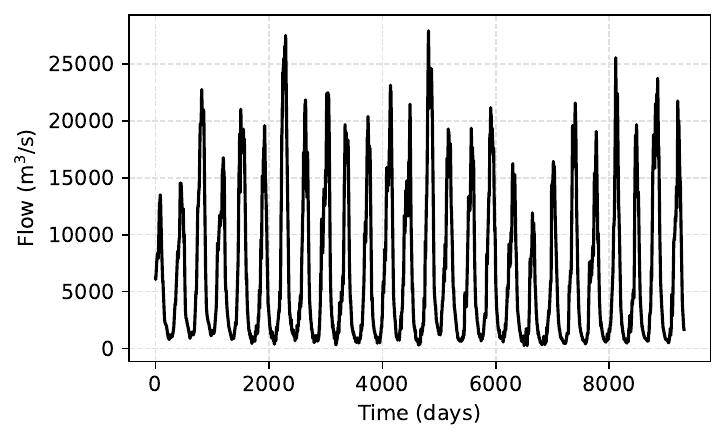}}
	\caption{ \label{f-O}Original flow time series data.}
\end{figure}

To evaluate the model, the historical flow record of a hydroelectric plant (Tucuru\'i) located in Brazil was used \cite{muniz2025time}. Figure \ref{f-O} shows the original time series. This record covers a time window from January 2, 1998 to July 9, 2023, which is a daily record comprising 9,321 recorded values. 

To perform a comparative analysis, we evaluate our algorithm against the 
neural basis expansion analysis with exogenous variables (NBEATSx) \cite{olivares2023neural},
long short-term memory (LSTM) \cite{10097568},
GRU \cite{9612011},
DeepAR \cite{salinas2020deepar},
temporal fusion transformer (TFT) \cite{10946853},
informer \cite{yang2022time}, 
patch time series transformer (PatchTST) \cite{10680084},
frequency enhanced decomposed transformer (FEDformer) \cite{ge2025advanced},
temporal convolutional network (TCN) \cite{10536729},
and TimesNet \cite{zuo2023ensemble}.

\section{Results and Discussion}  \label{4}

This section presents the results and comparisons of the use of the proposed CNN-TFT-SHAP-MHAW model.

\subsection{Fine Tuning}

\begin{figure}[htb!]
	\centering
	\setkeys{Gin}{width=0.49\textwidth}
	{\includegraphics{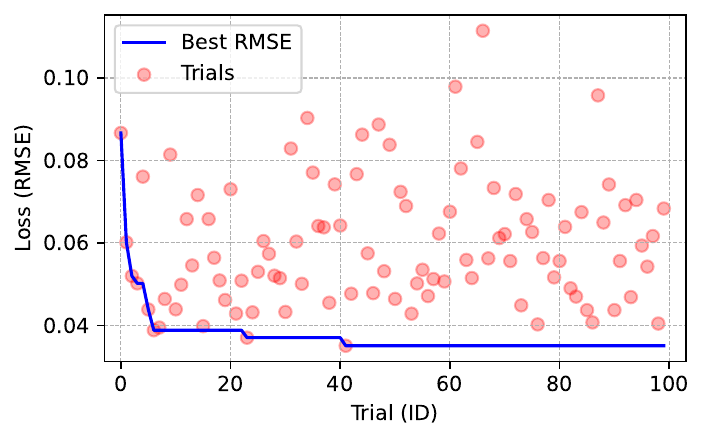}}
	\caption{ \label{f1}RMSE of the trial using Bayesian optimization for hypertuning.}
\end{figure}

The purpose of fine-tuning the model is to determine the optimal hyperparameters to be used in its structure, thus ensuring its best use. The Bayesian optimization hypertuning performs an interactive search to find the hyperparameters that minimize the loss function, which in this case is the RMSE. Figure \ref{f1} presents the loss (RMSE) of each trial in the experiments. The dots are the RMSE of each trial, and the blue line is the lowest loss seen by the Bayesian optimization method compared to earlier trials. Trial is a combination of hyperparameters considered by the Bayesian optimization method during hypertuning.

The results of the Bayesian optimization were 3 CNN layers, 4 heads for the attention mechanism, and 238 filters in the CNN using a kernel size of 4. Considering that all the hyperparameters were within the search space in the optimization process, the evaluation was adequate. 
Therefore, these values are used in our model. 

\begin{figure}[htb!]
	\centering
	\setkeys{Gin}{width=0.49\textwidth}
	{\includegraphics{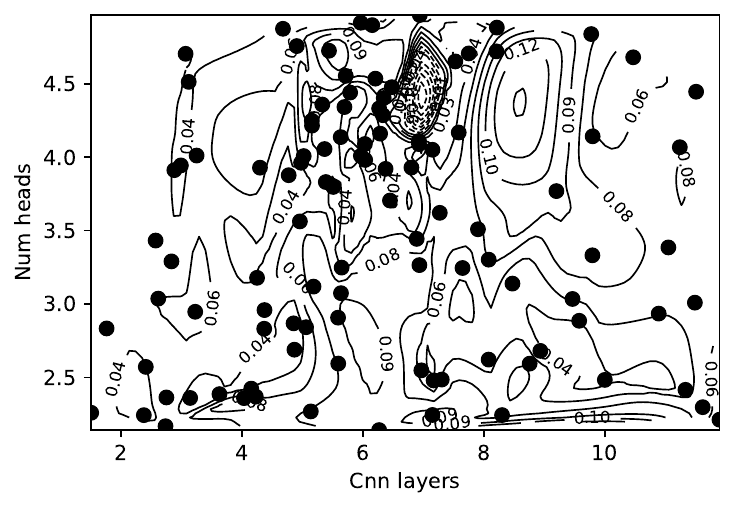}}
	\caption{ \label{f2}RMSE gradients of number of heads versus CNN layers.}
\end{figure}

Figure \ref{f2} shows the contour plot of RMSE values for a range of numbers of heads and CNN layers. The Results show that using 3 to 7 CNN layers, lower RMSE values are achieved, with a large variation in the number of heads, concerning the attention mechanism. 

\begin{figure}[htb!]
	\centering
	\setkeys{Gin}{width=0.49\textwidth}
    {\includegraphics{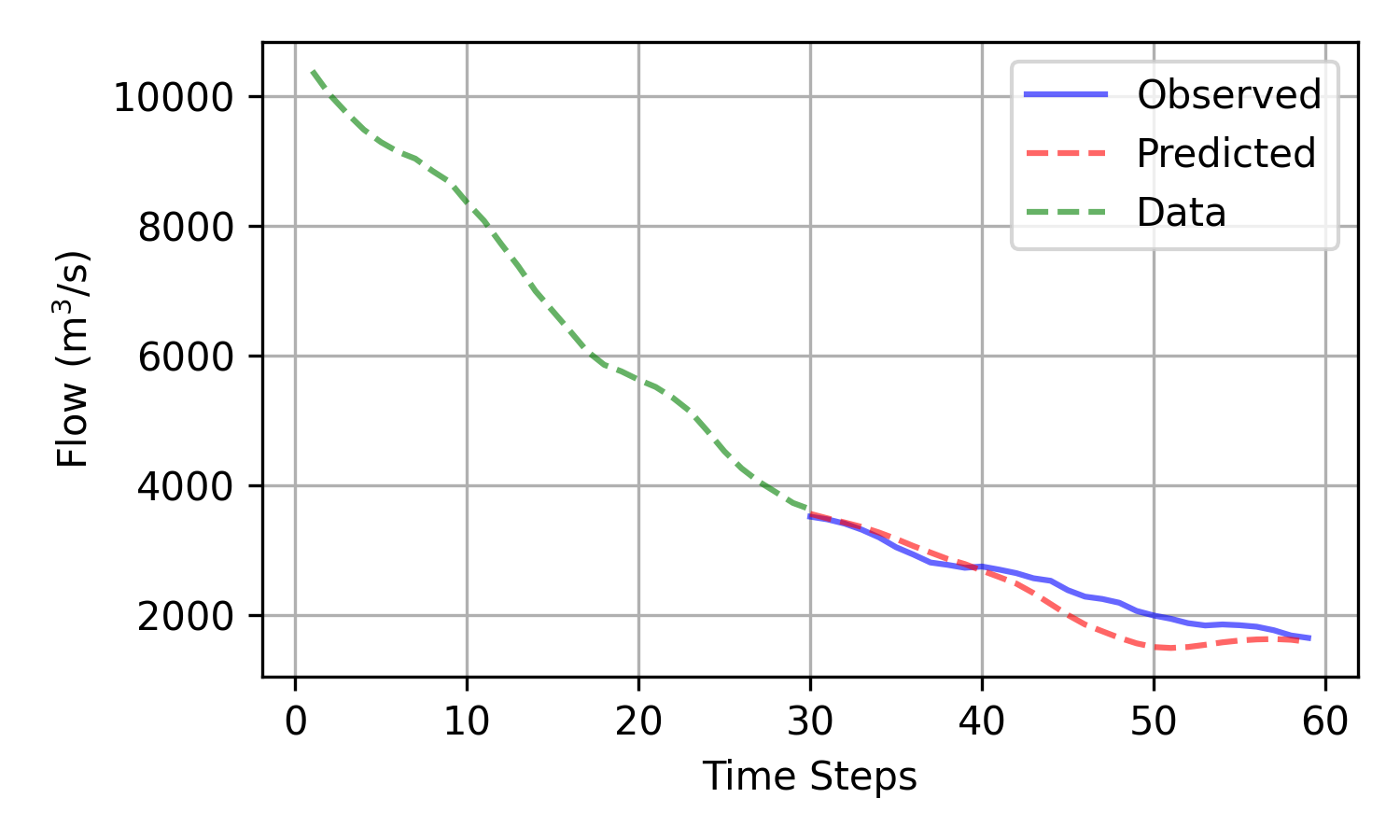}}
	\caption{ \label{f1aaa1}Example of original versus predicted time series by CNN-TFT-SHAP-MHAW.}
\end{figure}

Once these hyperparameters have been defined, the model is retrained to be applied to the original data; Figure \ref{f1aaa1} presents an example of prediction comparison to the original signal. These values of hyperparameters are also used for the statistical assessment, which is presented in the next subsection.

In this paper, we consider an input size equivalent to the forecasted horizon to train the model. Where the horizon refers to how far into the future you want to predict (time steps). Based on this hyperparameter setup, Table \ref{table:bm} presents a comparative analysis (benchmarking) to compare the proposed model (ours) to other well-established models.

\begin{table*}[htb!]
\caption{Comparative analysis of DL models.}
\label{table:bm}
\begin{tabular}{lllllllll}
\hline 
Model & Horizon & RMSE & MAE & MAPE & MSLE & Time \\ 
\hline

\multirow{4}{*}{NBEATSx} 
& 15 & 6.36$\times10^2$ & 5.82$\times10^2$ & 1.79$\times10^{-1}$ & 3.30$\times10^{-2}$ & 3.32$\times10^{-1}$ \\
& 30 & 4.55$\times10^2$ & 4.33$\times10^2$ & 1.54$\times10^{-1}$ & 2.23$\times10^{-2}$ & \textbf{2.37$\times10^{-1}$} \\
& 60 & 5.02$\times10^2$ & 4.31$\times10^2$ & 1.97$\times10^{-1}$ & 3.83$\times10^{-2}$ & 3.42$\times10^{-1}$ \\
& 120 & \textbf{5.40$\times10^2$} & \textbf{4.10$\times10^2$} & 4.93$\times10^{-1}$ & \textbf{2.29$\times10^{-1}$} & \textbf{3.51$\times10^{-1}$} \\

\cline{2-7}
\multirow{4}{*}{LSTM} 
& 15 & 1.39$\times10^3$ & 1.38$\times10^3$ & 4.04$\times10^{-1}$ & 1.18$\times10^{-1}$ & 2.91$\times10^{-1}$ \\
& 30 & 4.28$\times10^3$ & 4.25$\times10^3$ & 1.59 & 9.09$\times10^{-1}$ & 2.88$\times10^{-1}$ \\
& 60 & 1.22$\times10^4$ & 1.07$\times10^4$ & 6.62 & 3.62 & \textbf{3.32$\times10^{-1}$} \\
& 120 & 1.72$\times10^4$ & 1.70$\times10^4$ & 1.60$\times10^1$ & 7.57 & 4.05$\times10^{-1}$ \\
\cline{2-7}
\multirow{4}{*}{GRU} 
& 15 & 5.68$\times10^2$ & 4.51$\times10^2$ & 1.24$\times10^{-1}$ & 1.82$\times10^{-2}$ & \textbf{2.69$\times10^{-1}$} \\
& 30 & 1.44$\times10^3$ & 1.27$\times10^3$ & 4.46$\times10^{-1}$ & 1.47$\times10^{-1}$ & 2.85$\times10^{-1}$ \\
& 60 & 3.05$\times10^3$ & 2.86$\times10^3$ & 1.49 & 8.40$\times10^{-1}$ & 3.59$\times10^{-1}$ \\
& 120 & 1.96$\times10^4$ & 1.84$\times10^4$ & 1.78$\times10^1$ & 7.81 & 5.24$\times10^{-1}$ \\
\cline{2-7}
\multirow{4}{*}{DeepAR} 
& 15 & 3.55$\times10^3$ & 3.51$\times10^3$ & 1.00 & 6.39$\times10^1$ & 6.52$\times10^{-1}$ \\
& 30 & 3.01$\times10^3$ & 2.92$\times10^3$ & 1.00 & 6.08$\times10^1$ & 5.35$\times10^{-1}$ \\
& 60 & 2.36$\times10^3$ & 2.18$\times10^3$ & 1.00 & 5.68$\times10^1$ & 1.23 \\
& 120 & 1.76$\times10^3$ & 1.47$\times10^3$ & 1.00 & 4.99$\times10^1$ & 2.07 \\
\cline{2-7}
\multirow{4}{*}{TFT} 
& 15 & 4.85$\times10^2$ & 4.05$\times10^2$ & 1.24$\times10^{-1}$ & 2.04$\times10^{-2}$ & 2.03 \\
& 30 & 9.76$\times10^2$ & 8.12$\times10^2$ & 2.49$\times10^{-1}$ & 1.19$\times10^{-1}$ & 3.04 \\
& 60 & 2.18$\times10^3$ & 2.01$\times10^3$ & 9.18$\times10^{-1}$ & 6.47 & 5.47 \\
& 120 & 2.77$\times10^3$ & 2.59$\times10^3$ & 2.89 & 1.80 & 1.07$\times10^1$ \\
\cline{2-7}
\multirow{4}{*}{Informer} 
& 15 & 3.54$\times10^3$ & 3.51$\times10^3$ & 9.99$\times10^{-1}$ & 4.53$\times10^1$ & 1.50 \\
& 30 & 3.00$\times10^3$ & 2.91$\times10^3$ & 9.99$\times10^{-1}$ & 4.44$\times10^1$ & 1.40 \\
& 60 & 2.36$\times10^3$ & 2.18$\times10^3$ & 9.98$\times10^{-1}$ & 3.83$\times10^1$ & 3.75 \\
& 120 & 1.76$\times10^3$ & 1.47$\times10^3$ & 9.97$\times10^{-1}$ & 3.20$\times10^1$ & 6.93 \\
\cline{2-7}
\multirow{4}{*}{PatchTST} 
& 15 & 5.95$\times10^2$ & 5.16$\times10^2$ & 1.57$\times10^{-1}$ & 2.76$\times10^{-2}$ & 3.80$\times10^{-1}$ \\
& 30 & 4.47$\times10^2$ & 3.78$\times10^2$ & 1.50$\times10^{-1}$ & 3.83$\times10^{-2}$ & 4.28$\times10^{-1}$ \\
& 60 & 1.76$\times10^3$ & 1.53$\times10^3$ & 9.39$\times10^{-1}$ & 1.35 & 5.85$\times10^{-1}$ \\
& 120 & 1.26$\times10^3$ & 9.98$\times10^2$ & 9.29$\times10^{-1}$ & 1.21 & 1.03 \\
\cline{2-7}
\multirow{4}{*}{FEDformer} 
& 15 & 2.89$\times10^3$ & 2.84$\times10^3$ & 8.49$\times10^{-1}$ & 3.86$\times10^{-1}$ & 2.73 \\
& 30 & 5.41$\times10^3$ & 5.35$\times10^3$ & 2.02 & 1.22 & 3.42 \\
& 60 & 1.11$\times10^4$ & 1.11$\times10^4$ & 6.12 & 3.72 & 1.84$\times10^1$ \\
& 120 & 1.35$\times10^4$ & 1.35$\times10^4$ & 1.33$\times10^1$ & 6.61 & 4.74$\times10^1$ \\
\cline{2-7}
\multirow{4}{*}{TCN} 
& 15 & 1.69$\times10^3$ & 1.45$\times10^3$ & 4.28$\times10^{-1}$ & 2.44$\times10^{-1}$ & 3.98$\times10^{-1}$ \\
& 30 & 2.98$\times10^3$ & 2.39$\times10^3$ & 8.91$\times10^{-1}$ & 6.21$\times10^{-1}$ & 2.64$\times10^{-1}$ \\
& 60 & 6.00$\times10^3$ & 5.03$\times10^3$ & 2.84 & 1.77 & 4.24$\times10^{-1}$ \\
& 120 & 6.71$\times10^3$ & 5.44$\times10^3$ & 5.38 & 3.20 & 5.20$\times10^{-1}$ \\
\cline{2-7}
\multirow{4}{*}{TimesNet} 
& 15 & 1.27$\times10^3$ & 1.01$\times10^3$ & 3.16$\times10^{-1}$ & 1.03$\times10^{-1}$ & 3.73 \\
& 30 & 1.99$\times10^3$ & 1.67$\times10^3$ & 6.04$\times10^{-1}$ & 8.05$\times10^{-1}$ & 8.38 \\
& 60 & 4.20$\times10^3$ & 3.34$\times10^3$ & 1.64 & 8.97$\times10^{-1}$ & 1.66$\times10^1$ \\
& 120 & 3.88$\times10^3$ & 3.38$\times10^3$ & 3.03 & 1.79 & 5.47$\times10^1$ \\
\cline{2-7}
\multirow{4}{*}{Ours} 
& 15 & \textbf{9.28$\times10^1$} & \textbf{7.37$\times10^1$} & \textbf{2.20$\times10^{-2}$} & \textbf{6.88$\times10^{-4}$} & 1.90$\times10^1$ \\ 
& 30 & \textbf{2.01$\times10^2$} & \textbf{1.76$\times10^2$} & \textbf{7.40$\times10^{-2}$} & \textbf{8.10$\times10^{-3}$} & 2.09$\times10^1$ \\
& 60 & \textbf{2.92$\times10^2$} & \textbf{2.23$\times10^2$} & \textbf{1.37$\times10^{-1}$} & \textbf{3.89$\times10^{-2}$} & 2.16$\times10^1$ \\
& 120 & 5.45$\times10^2$ & 4.69$\times10^2$ & 6.46$\times10^{-1}$ & 5.69$\times10^{-1}$ & 1.81$\times10^1$ \\
\cline{2-7}
\hline
\multicolumn{3}{l}{Best results for each horizon in bold.}\\
\end{tabular} \centering
\end{table*}

The results are computed considering the average of RMSE, MAE, MAPE, and MSLE, where the horizon is the predicted time steps. 
Our model had promising results, especially for horizons equal to 15, 30, and 60 steps ahead (being the best model). When longer horizons were evaluated, the best results were given by NHITS and NBEATSx. For a horizon equal to 120 steps ahead, 
the NBEATSx had 0.92\% better results regarding RMSE. In this case, our model was the second-best among all the models compared.

\subsection{Statistical Assessment}

After hypertuning, we select the values of each hyperparameter, then we use these values to compute experiments (runs) with initialization of the random weights. 
We performed the analysis on 50 runs. Each run is computed by training the model up to 100 epochs and evaluating it, using 80\% of the data for training and 20\% for testing.

Table \ref{tab:extended_stats_eng} shows the results of the statistical analysis of the RMSE, MAE, MAPE, MSLE for 50 runs, where the mean, median, standard deviation, first quartile, third quartile, IQR, Skewness, and Kurtosis of the 50 error measures are computed. The statistical analysis aims to assess the variability of the model when several simulations are carried out. 

\begin{table}[!ht]
\centering
\caption{Statistical RMSE Results over 50 Runs.}
\begin{tabular}{lccccccccccc}
\hline
Statistic & RMSE & MAE & MAPE & MSLE \\
\hline

Mean & 5.22$\times10^{2}$ & 4.19$\times10^{2}$ & 1.37$\times10^{-1}$ & 3.08$\times10^{-2}$ \\
Std & 2.55$\times10^{2}$ & 1.57$\times10^{2}$ & 5.84$\times10^{-2}$ & 2.69$\times10^{-2}$ \\
Min & 2.35$\times10^{2}$ & 1.91$\times10^{2}$ & 5.94$\times10^{-2}$ & 5.76$\times10^{-3}$ \\
Max & 1.25$\times10^{3}$ & 8.68$\times10^{2}$ & 3.07$\times10^{-1}$ & 1.13$\times10^{-1}$ \\
Median & 4.55$\times10^{2}$ & 3.81$\times10^{2}$ & 1.23$\times10^{-1}$ & 2.17$\times10^{-2}$ \\
Q1 (25\%) & 3.70$\times10^{2}$ & 3.09$\times10^{2}$ & 9.77$\times10^{-2}$ & 1.43$\times10^{-2}$ \\
Q3 (75\%) & 5.49$\times10^{2}$ & 5.11$\times10^{2}$ & 1.63$\times10^{-1}$ & 3.55$\times10^{-2}$ \\
Range & 1.01$\times10^{3}$ & 6.77$\times10^{2}$ & 2.47$\times10^{-1}$ & 1.07$\times10^{-1}$ \\
IQR & 1.80$\times10^{2}$ & 2.03$\times10^{2}$ & 6.52$\times10^{-2}$ & 2.13$\times10^{-2}$ \\
Skewness & 1.73 & 1.03 & 1.27 & 1.85 \\
Kurtosis & 2.09 & 5.57$\times10^{-1}$ & 1.03 & 2.47 \\
\hline
\end{tabular}
\label{tab:extended_stats_eng}
\end{table}
   
Over 100 independent trials, the model's errors averaged an RMSE of 521.5, MAE of 419.5, MAPE of 13.7\%, and MSLE of 0.0308, demonstrating moderate performance. The RMSE and MAE error distributions are positively skewed (skewness $>$1 across all metrics), reflecting occasional runs with particularly large mistakes; the worst RMSE reached 1,246.5, and the highest MAPE was 30.7\%. 

\begin{figure}[htb!]
	\centering
	\setkeys{Gin}{width=0.49\textwidth}
	{\includegraphics{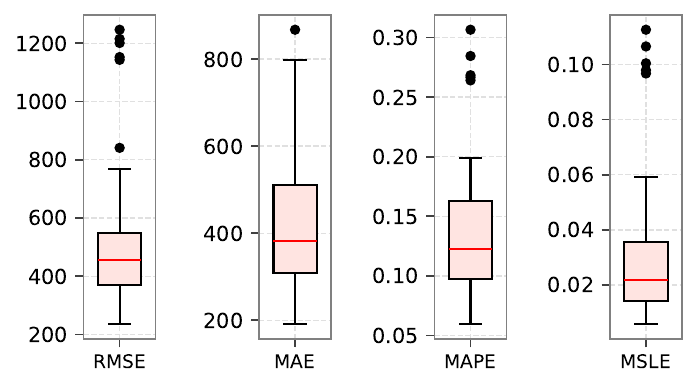}}
	\caption{ \label{fS}Results of 50 runs of our proposed model.}
\end{figure}

Figure \ref{fS} shows the box plot of the error measures of the 50 runs. The red line (inside the rectangle) is the median, and the pink rectangle is the IQR range (from Q1 to Q3). Here, the outliers are the black individual dots beyond the whiskers (Q1-(1.5*IQR) or Q3+(1.5*IQR)).

The IQRs from Figure \ref{fS} further illustrate consistency bounds: 50\% of RMSE values fell between 369.5 and 549.3, while the MAE IQR spanned 308.7 to 511.5. Altogether, these statistics suggest a model that performs well on average but exhibits sporadic outliers, indicating opportunities for improved stability and robustness.

Considering the results of low variability concerning the metrics evaluated, with few outliers in relation to the experiments carried out, the proposed model proved to be robust to random initialization of the network, making it suitable for the application addressed in this work.

\subsection{Explainable Results}

\begin{figure*}[htb!]
  \centering
  \setkeys{Gin}{width=0.95\textwidth}
  \begin{subfigure}{0.49\textwidth}
    \includegraphics{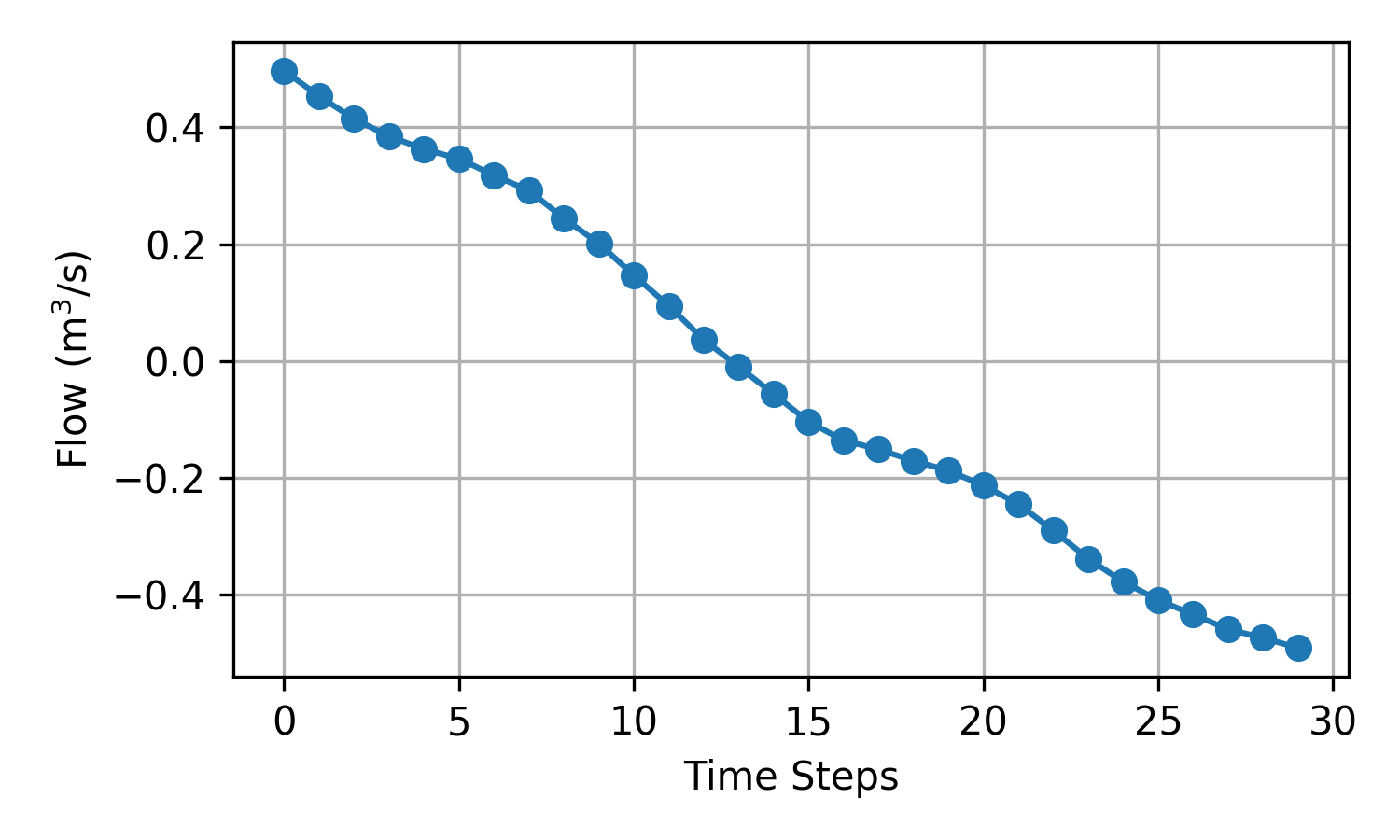}
    \caption{Observed time series.}
    \label{fig:subfig1}
  \end{subfigure}
  \hfill
  \begin{subfigure}{0.49\textwidth}
    \includegraphics{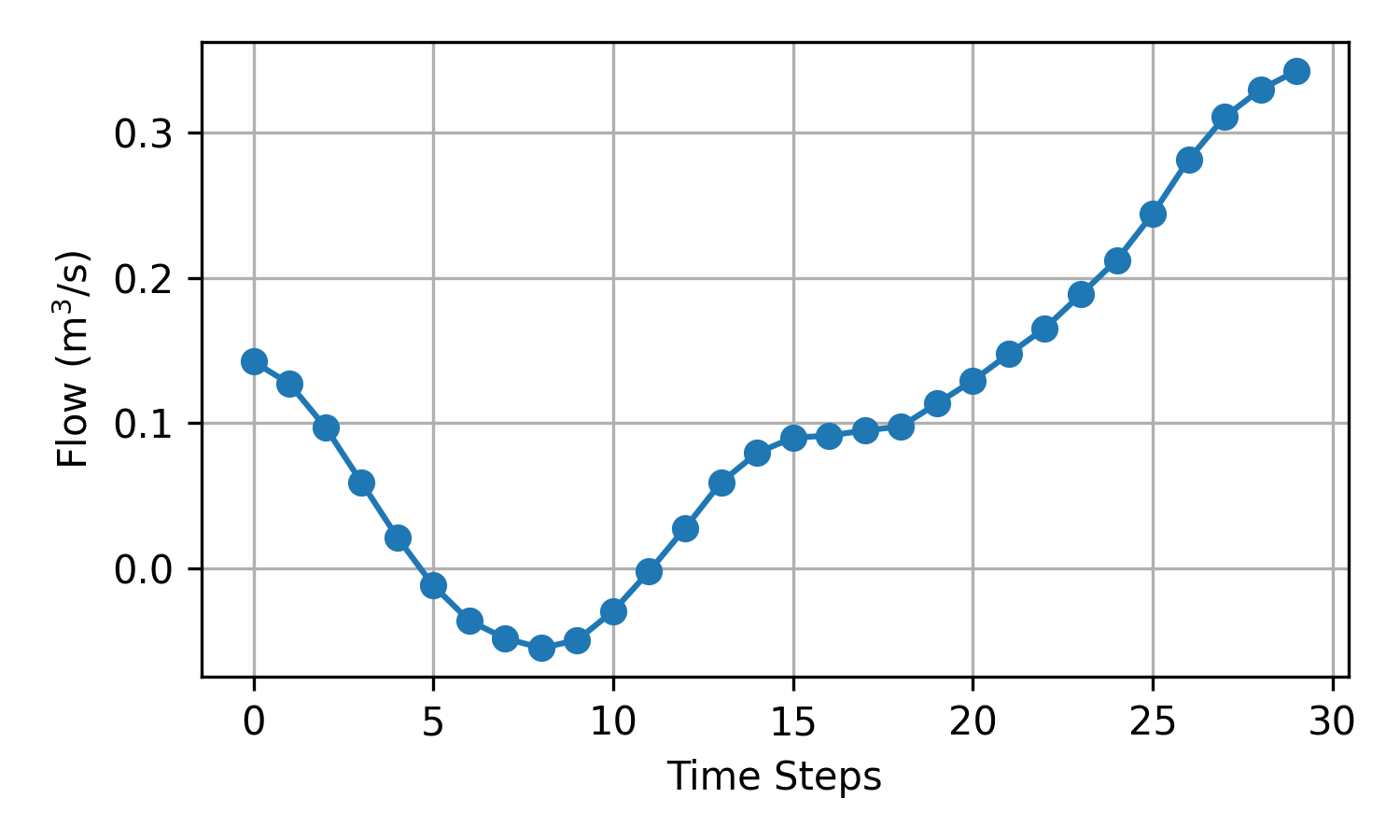}
    \caption{Observed time series.}
    \label{fig:subfig2}
  \end{subfigure}

  \begin{subfigure}{0.49\textwidth}
    \includegraphics{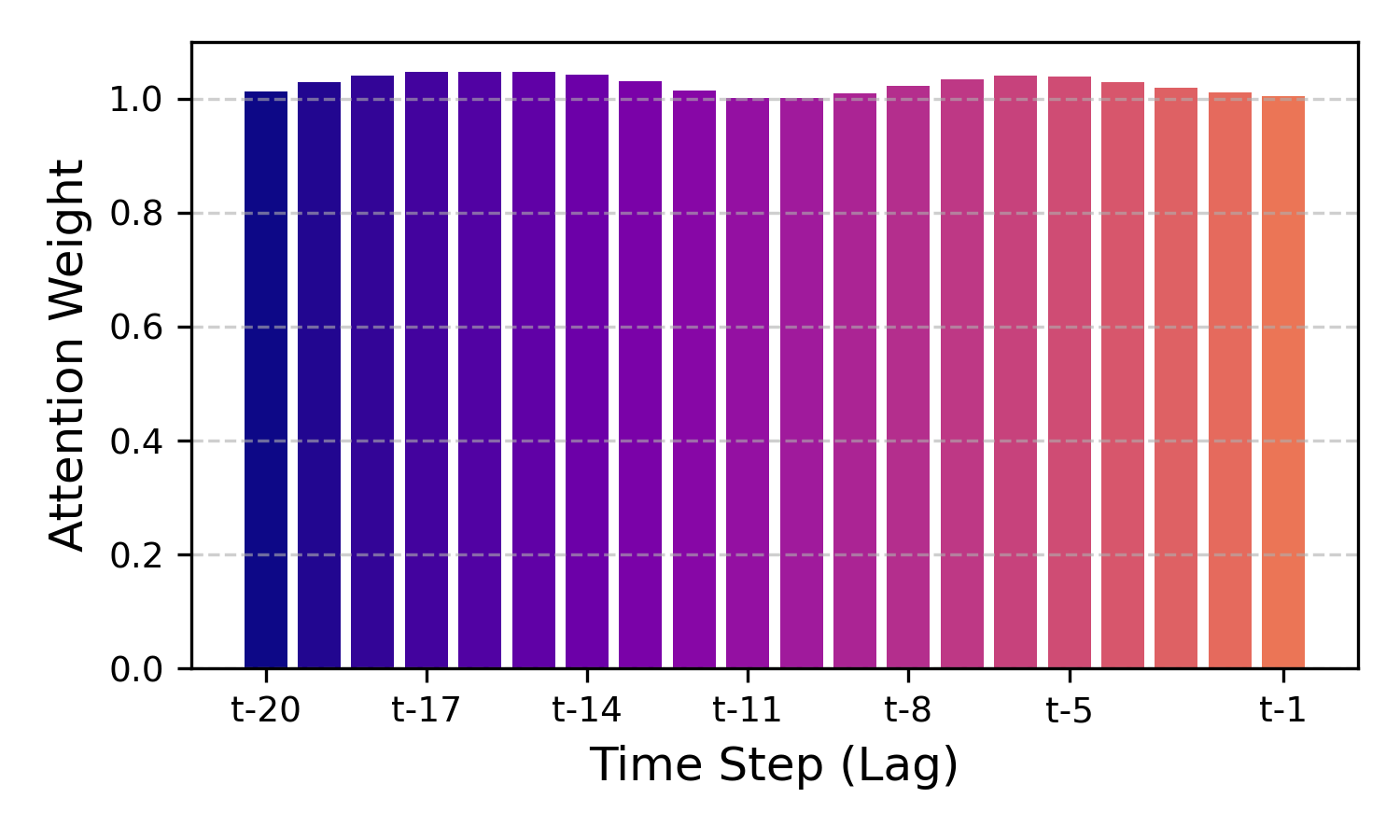}
    \caption{Mean attention weights.}
    \label{fig:subfig3}
  \end{subfigure}
  \hfill
  \begin{subfigure}{0.49\textwidth}
    \includegraphics{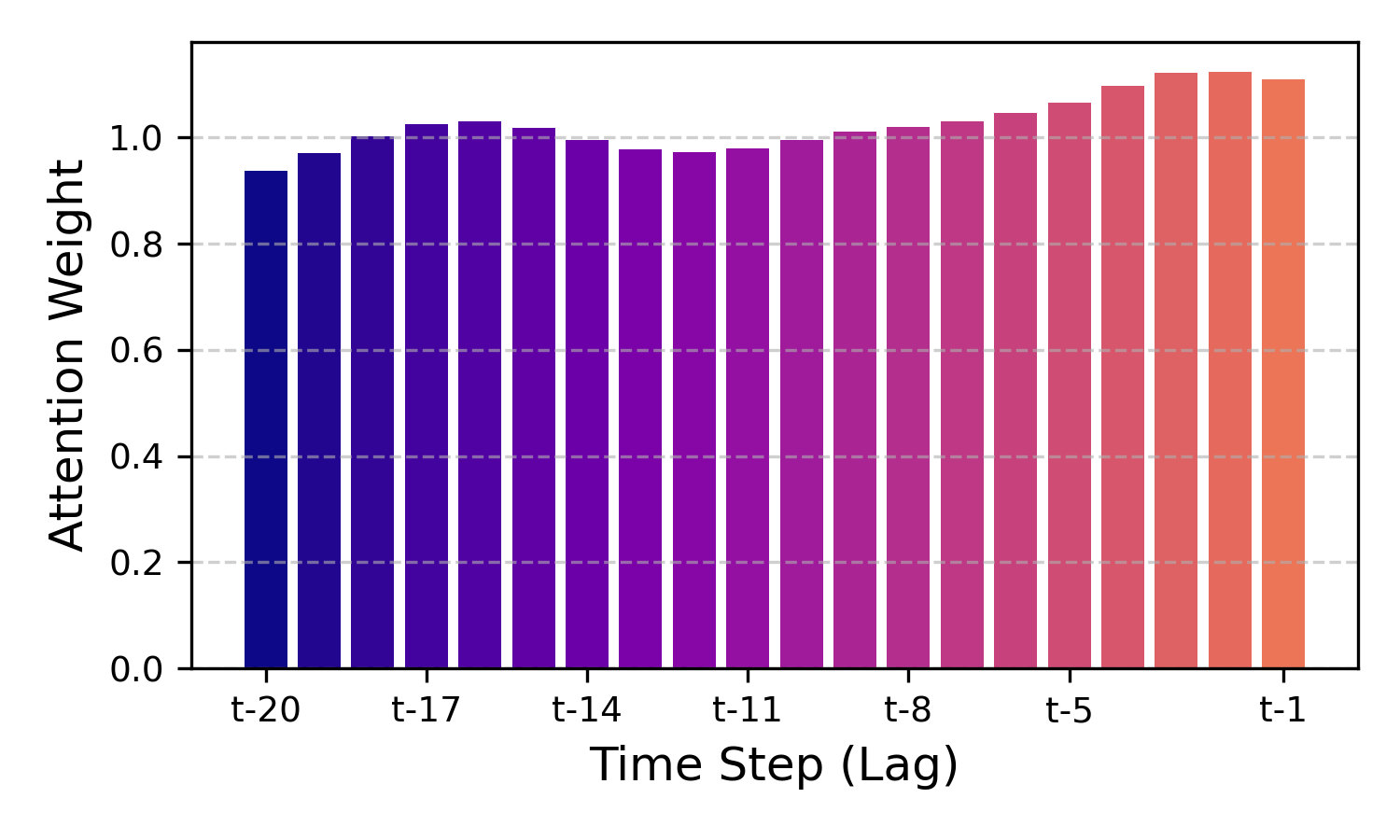}
    \caption{Mean attention weights}
    \label{fig:subfig4}
  \end{subfigure}

  \begin{subfigure}{0.49\textwidth}
    \includegraphics{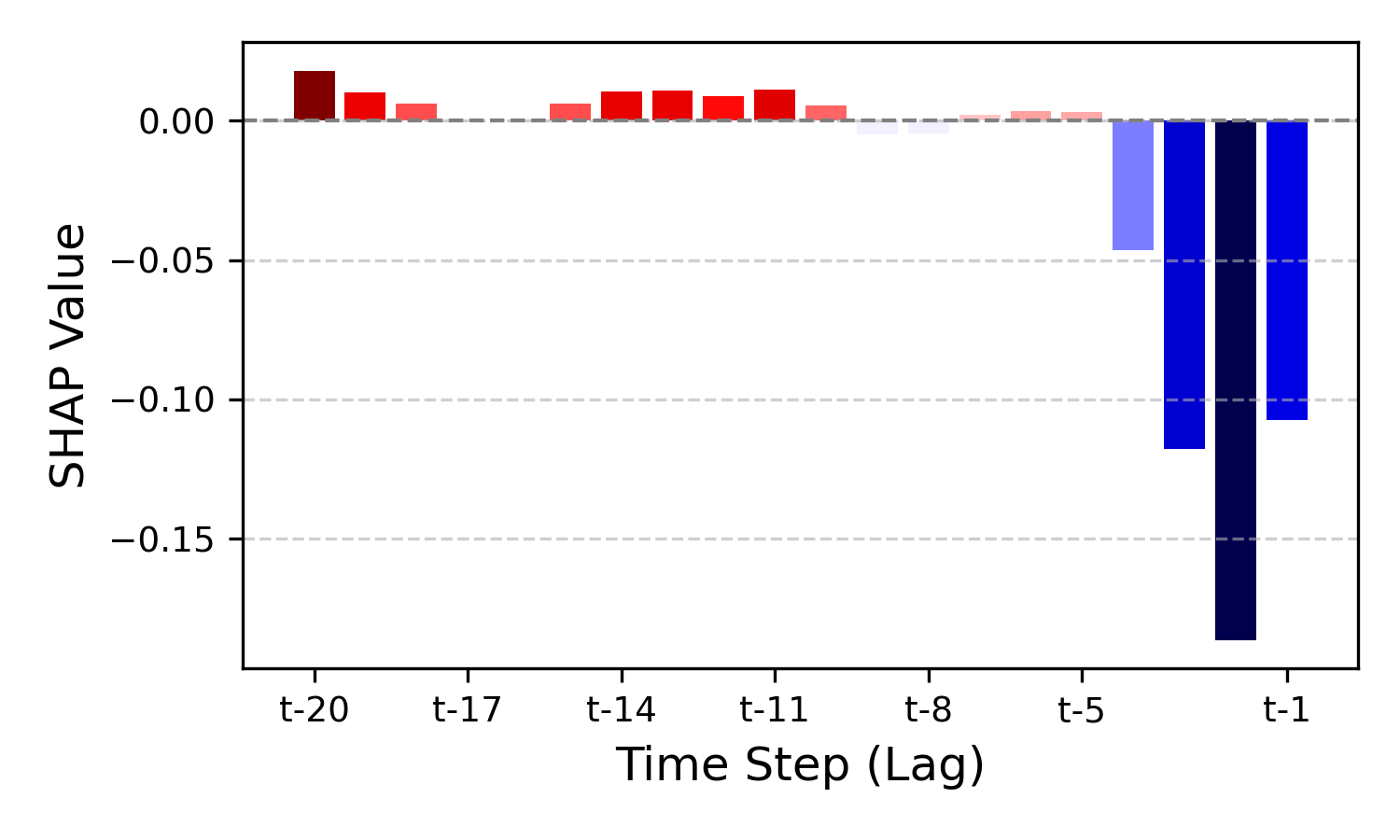}
    \caption{SHAP value for a batch sample.}
    \label{fig:subfig5}
  \end{subfigure}
  \hfill
  \begin{subfigure}{0.49\textwidth}
    \includegraphics{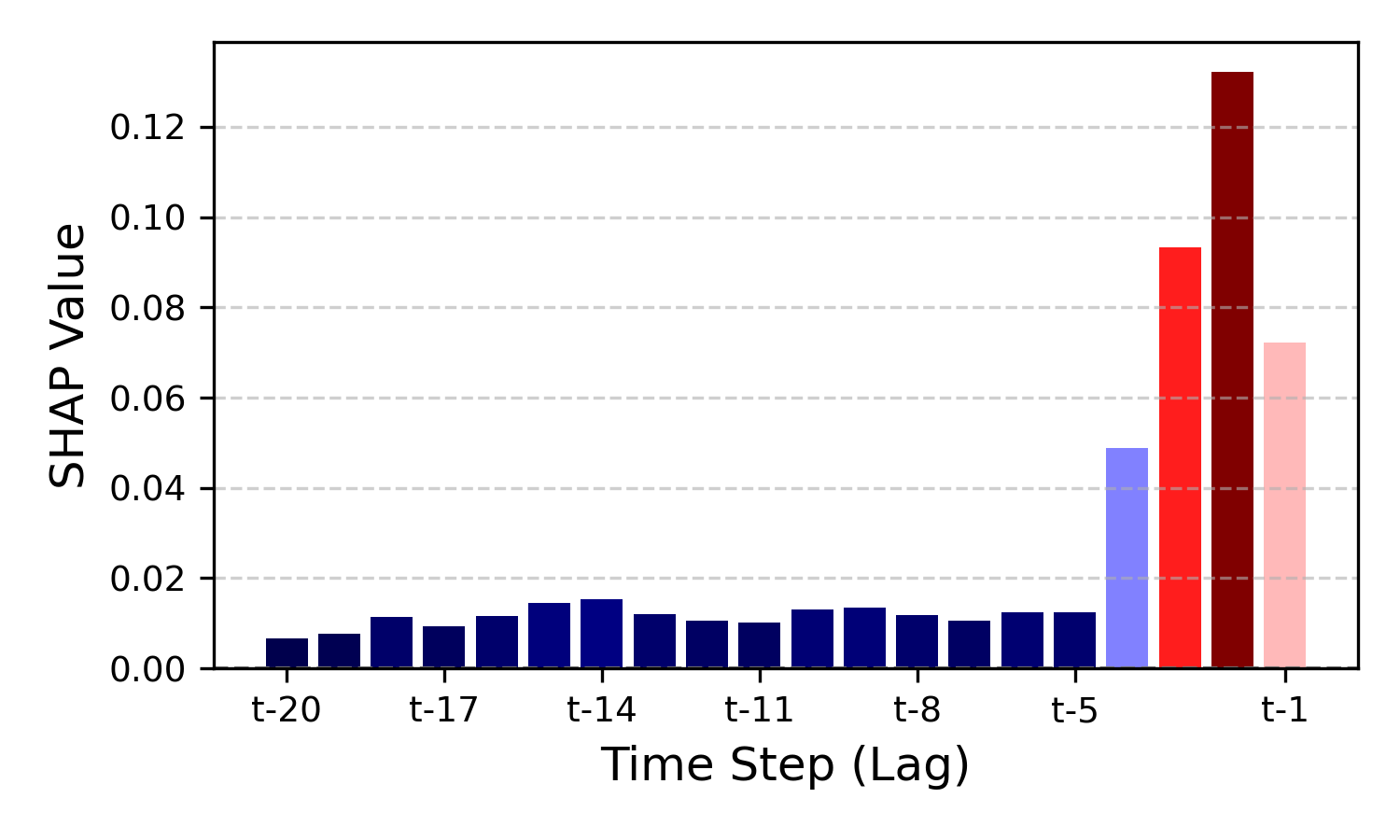}
    \caption{SHAP value for a batch sample.}
    \label{fig:subfig6}
  \end{subfigure}

  \begin{subfigure}{0.49\textwidth}
    \includegraphics{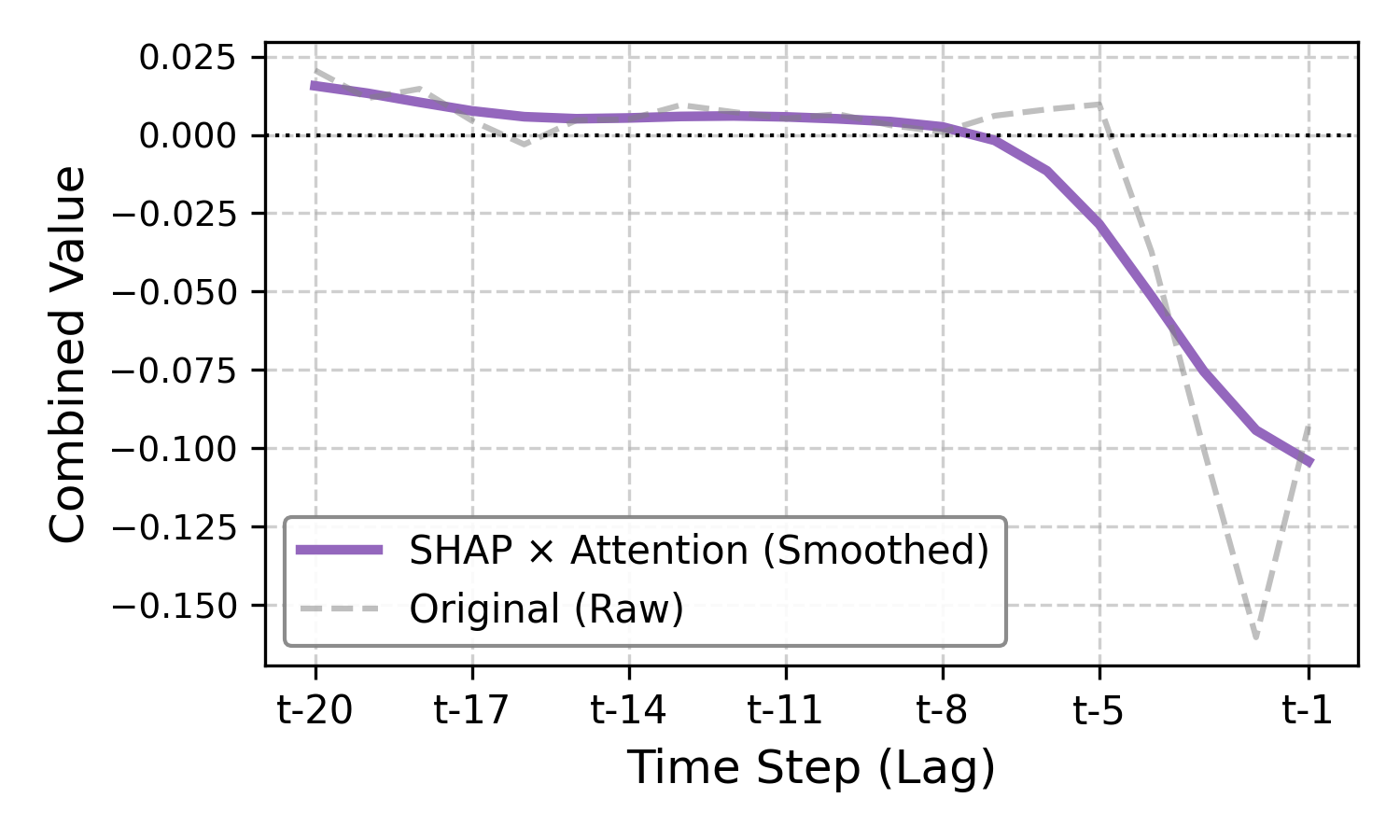}
    \caption{Combined influence map: SHAP $\times$ attention.}
    \label{fig:subfig7}
  \end{subfigure}
  \hfill
  \begin{subfigure}{0.49\textwidth}
    \includegraphics{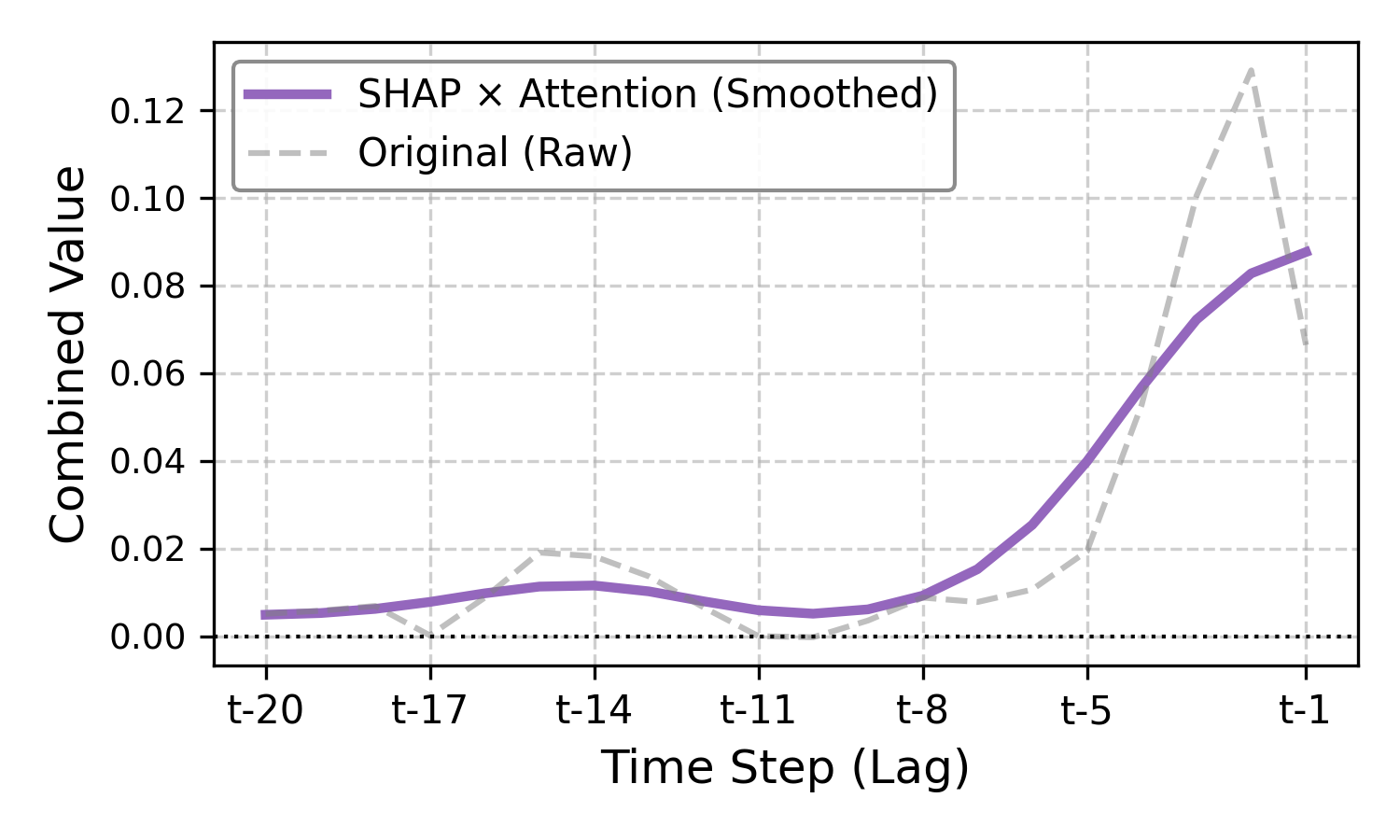}
    \caption{Combined influence map: SHAP $\times$ attention.}
    \label{fig:subfig8}
  \end{subfigure}

  \caption{Explainable results for a given time series.}
  \label{fig:all_subfigs}
\end{figure*}

This subsection presents the effect of using explainable methods in our model. Figure \ref{fig:all_subfigs} (a) and (b) show the observed time series, Figure \ref{fig:all_subfigs} (c) and (d) show the multi-head attention weights, Figure \ref{fig:all_subfigs} (e) and (f) show the SHAP values, and the Figure \ref{fig:all_subfigs} (g) and (h) show the combined SHAP $\times$ attention values.

 These weights reflect the distribution of attention across the input window during prediction. 
For the attention and SHAP values, those values that are far away from the current sample and near the edge of the window (e.g., t-29 in a window size of 30) they are showing erratic values because of the influence from the border values. Due to this, we do not consider these last values.

In Figure \ref{fig:all_subfigs} (c) and (d) it is observed the influence of seasonal variability of the time series on the attention weights. Figure \ref{fig:all_subfigs} (e) and (f) present the SHAP values assigned to each time lag. Positive values (in red) indicate a contribution to increase the forecast, while negative values (in blue) indicate a contribution to decrease it. Quantitative analysis reveals that the majority of the explanatory impact is concentrated in the more recent values of the input window (e.g., $t\text{-}10$ to $t\text{-}1$), which accounts for approximately 85.37\% of the total SHAP magnitude.


Figure \ref{fig:all_subfigs} (g) and (h) present the combined influence map, obtained by computing the element-wise product between SHAP values and attention weights for each time lag. 
The purple line represents the smoothed influence signal after applying a one-dimensional Gaussian filter, while the gray dashed line corresponds to the raw product. The smoothed curve improves interpretability by reducing local noise and emphasizing overall trends. This combination highlights the time steps that are both attended to by the model and have a meaningful influence on the final prediction.

The plot reveals that the most influential regions are concentrated near the end of the input window, particularly between $t\text{-}6$ and $t\text{-}1$, where the influence varies sharply. This aligns with the expectation that recent lags are often more informative in time series forecasting. However, the curve also shows moderate combined relevance in earlier parts of the window, reflecting the presence of temporal dependencies beyond the immediate past. 

The use of SHAP$\times$Attention thus provides a more reliable and interpretable measure of temporal influence than either method alone where SHAP values quantify the actual contribution of each time step to the model's prediction, indicating how each lag increases or decreases the forecast, while attention weights reflect where the model is focusing during the decision process, regardless of whether that focus has a positive, negative, or neutral impact.

\section{Conclusion} \label{5}

This study introduces CNN-TFT-SHAP-MHAW, a hybrid deep neural network architecture that integrates convolutions and a multi-head attention mechanism to enhance multivariate time series forecasting. By using the pattern recognition of CNNs alongside the dynamic attention mechanisms of transformers, the proposed model can effectively capture both localized temporal features and long-range dependencies. The convolutional component reduces input noise and dimensionality, high-level features that enhance the representation power of the subsequent transformer layers. The TFT backbone applies multi-head attention and adaptive gating mechanisms to emphasize relevant covariates and temporally important patterns, enabling more accurate forecasting.

Experiments were conducted 
considering a hydroelectric natural flow 
time series data, demonstrating the superior performance of CNN-TFT-SHAP-MHAW when compared to several state-of-the-art DL models. The model achieves an MAPE as low as 2.2\%, underscoring its effectiveness and robustness across different time series forecasting scenarios. These findings validate the hybrid model's ability to generalize across domains with varying temporal structures and feature complexities.

By merging attention weights (model focus) and SHAP values (causal contribution), the resulting influence map improved interpretability, allowing for a more trustworthy understanding of which past inputs drive the forecast. The integration of SHAP values with multi-head attention visualization provides more interpretable results to assist the decision-making process.

The implications of this research extend to numerous real-world applications that demand high-fidelity temporal predictions. The CNN-TFT-SHAP-MHAW model facilitates the deployment of more accurate and interpretable models in decision-making processes. Future work may explore the integration of external knowledge sources, enhanced interpretability modules, and real-time deployment frameworks to further extend the utility and adaptability of the CNN-TFT-SHAP-MHAW architecture.


\bibliography{ref}
\bibliographystyle{IEEEtran}

\newpage

\begin{IEEEbiography}[{\includegraphics[width=1in,height=1.25in,clip,keepaspectratio]{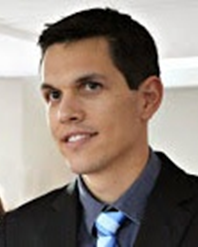}}]{Stefano Frizzo Stefenon}
received the B.E. and M.E. degrees in electrical engineering from the Regional University of Blumenau, Brazil, in 2012 and 2015, respectively, and his Ph.D. degree in electrical engineering from the State University of Santa Catarina, Brazil, in 2021. He was a Post-Doctoral Fellow researcher with the Faculty of Engineering and Applied Science, University of Regina, Canada.
He is currently a Professor with the Lisbon School of Engineering (ISEL), Polytechnic University of Lisbon, Portugal.
His research interests include machine learning for time series forecasting and fault prediction.
\end{IEEEbiography}

\vspace{-1pt}

\begin{IEEEbiography}[{\includegraphics[width=1in,height=1.25in,clip,keepaspectratio]{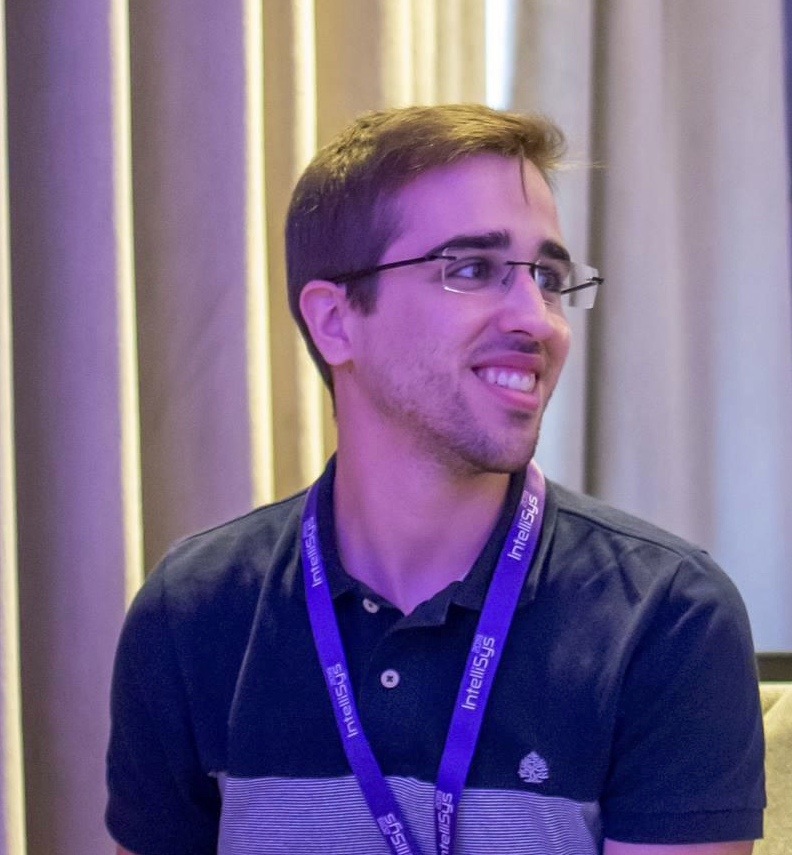}}]{Jo\~ao Pedro Matos-Carvalho}
received the M.Sc. (Hons.) and Ph.D. degrees in electrical and computer engineering from FCT NOVA, Portugal, in 2017 and 2021, respectively. He is currently an Assistant Professor at the Dept. of Informatics of the Faculty of Sciences of the University of Lisbon. Since 2025, he has been an Integrated Member of LASIGE and a collaborator of the Center of Technology and Systems (CTS), UNINOVA, and COPELABS. He won the highly competitive Scientific Employment Stimulus (CEEC institutional) FCT grant in 2021. 
He has published more than 50 papers in international journals and international conferences in the fields of remote sensing, pattern recognition, machine learning, sensor networks, and signal processing, with significant recognition and impact in the research community.
\end{IEEEbiography}

\vspace{-1pt}

\begin{IEEEbiography}[{\includegraphics[width=1in,height=1.25in,clip,keepaspectratio]{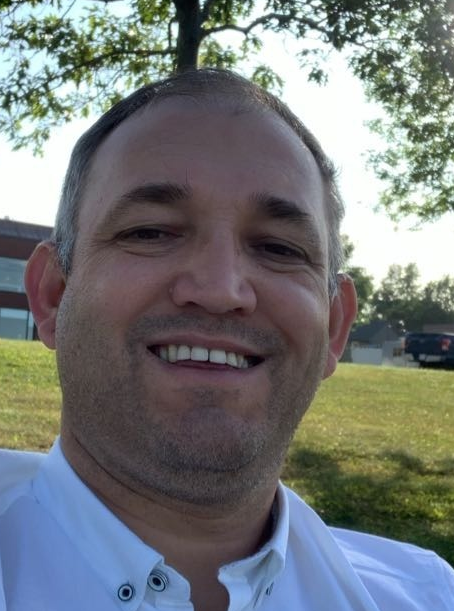}}]{Valderi Reis Quietinho Leithardt}
(Senior Member, IEEE) received the Ph.D. degree in computer science from INF-UFRGS, Brazil, in 2015. 
He is currently a Professor with the ISCTE - Instituto Universitario de Lisboa and a Researcher integrated with the Istar-Information Sciences, Technologies and Architecture Research Centre (ISTA), and Research Group Software Systems Engineering. 

He also has collaborations with Brazilian and Spanish researchers. His research interests include distributed systems with a focus on data privacy, communication, and programming protocols, involving scenarios and applications for Internet of Things, smart cities, big data, and cloud computing.
\end{IEEEbiography}

\vspace{-1pt}

\begin{IEEEbiography}[{\includegraphics[width=1in,height=1.25in,clip,keepaspectratio]{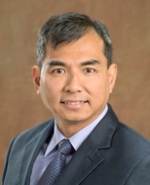}}]{Kin-Choong Yow}
(Senior Member, IEEE) obtained his B.E. (Elect) with  1st  Class  Honours from the  National  University of Singapore in 1993, and his Ph.D. from Cambridge University, UK in 1998. 

He joined the University of Regina (UofR) in September 2018, where he is presently a Full Professor in the Faculty of Engineering and  Applied  Science, Canada. 
Prior to joining UofR, he was an Associate Professor at the Gwangju Institute of Science and Technology (GIST),  Republic of  Korea  (2013-2018), Professor at the Shenzhen Institutes of Advanced Technology (SIAT), China (2012-2013), and Associate Professor at the Nanyang Technological University (NTU), Singapore (1998-2013). 
From 1999-2005, he served as the Sub-Dean of Computer Engineering at NTU, and from 2006-2008, he served as the Associate Dean of Admissions at NTU. 
\end{IEEEbiography}


\vfill

\end{document}